\documentclass[journal]{IEEEtran}

\usepackage{etex}
\reserveinserts{1000}
\usepackage{commath}
\usepackage{cite}
\usepackage{amsmath,amssymb,amsfonts}
\usepackage{graphicx}
\usepackage{textcomp}
\usepackage[english]{babel}
\usepackage[normalem]{ulem}
\usepackage{graphicx,epstopdf}
\usepackage[table]{xcolor}
\usepackage{booktabs}
\usepackage{latexsym}
\usepackage{eurosym}
\usepackage{mathtools}
\usepackage{lscape}
\usepackage{xcolor}
\usepackage{lipsum}
\usepackage{optidef}
\usepackage{multirow}
\usepackage{siunitx}
\usepackage{subcaption}
\usepackage{algorithm}
\usepackage{algorithmic}
\usepackage{pifont}
\usepackage{hyperref}

\usepackage{amsthm}
\usepackage{tcolorbox}

\usepackage{hyperref}
\hypersetup{hidelinks}
\usepackage{fancyhdr}
\usepackage{lipsum} % For dummy text

\begin{document}
\IEEEoverridecommandlockouts
\newcommand{\cmark}{\ding{51}}%
\newcommand{\xmark}{\ding{55}}%
\title{Hybrid LLM-DDQN based Joint Optimization of V2I Communication and Autonomous Driving \vspace{-1mm}}

\author{Zijiang~Yan, %~\IEEEmembership{Student~Member,~IEEE,}
        Hao~Zhou, %~\IEEEmembership{Member,~IEEE,}
        Hina~Tabassum,~\IEEEmembership{Senior~Member,~IEEE,} % <-this % stops a space
        and Xue~Liu,~\IEEEmembership{Fellow,~IEEE}
        \vspace{-10pt}
\thanks{Z.Yan and H.Tabassum are with the Department of Electrical Engineering and Computer Science, York University, Toronto,
ON, M3J 1P3, Canada. e-mail: \{zjiyan,hinat\}@yorku.ca; H. Zhou and X. Liu are with the Department of Computer Science, McGill University, Montréal, QC, H3A 0G4, Canada. e-mail: hao.zhou4@mail.mcgill.ca, xueliu@cs.mcgill.ca.
{For a more comprehensive understanding of our work, we provide supplementary experiment results and discussions in reference \cite{supple}.}}% <-this % stops a space
}

\maketitle
  
\begin{abstract}
Large language models (LLMs) have received considerable interest recently due to their outstanding reasoning and comprehension capabilities.
This work explores applying LLMs to vehicular networks, aiming to jointly optimize vehicle-to-infrastructure (V2I) communications and autonomous driving (AD) policies. 
We deploy LLMs for AD decision-making to maximize traffic flow and avoid collisions for road safety, and a double deep Q-learning algorithm (DDQN) is used for V2I optimization to maximize the received data rate and reduce frequent handovers.
{In particular, for LLM-enabled AD, we employ the Euclidean distance to identify previously explored AD experiences, and then LLMs can learn from past good and bad decisions for further improvement.    
Then, LLM-based AD decisions will become part of states in V2I problems, and DDQN will optimize the V2I decisions accordingly.}
After that, the AD and V2I decisions are iteratively optimized until convergence. 
Such an iterative optimization approach can better explore the interactions between LLMs and conventional reinforcement learning techniques, revealing the potential of using LLMs for network optimization and management.
Finally, the simulations demonstrate that our proposed hybrid LLM-DDQN approach outperforms the conventional DDQN algorithm, showing faster convergence and higher average rewards.
\end{abstract}

% Note that keywords are not normally used for pee rreview papers.
\begin{IEEEkeywords}
% IEEE, IEEEtran, journal, \LaTeX, paper, template.
Autonomous driving, Large language model, Vehicular networks, Network optimization
\end{IEEEkeywords}

\IEEEpeerreviewmaketitle

\raggedbottom

\section{Introduction}

%\textcolor{red}{Significance of LLM}
%\textcolor{red}{LLM for (RL)}
%\textcolor{red}{Why LLM for transportation and DQN for Telecom}
\thispagestyle{fancy}

Large Language Models (LLMs) are emerging as a promising technique  to address a wide range of downstream tasks such as classification, prediction, and optimization \cite{zhou2024large}.
Conventional algorithms, such as convex optimization and reinforcement learning (RL), experience scalability issues.
For example, RL usually experiences a large number of iterations and well-known low sampling efficiency issues, and convex optimization requires dedicated problem transformation for convexity. 
By contrast, existing studies have shown that LLM-inspired optimization, i.e., in-context learning-based approaches, have several unique advantages: 1) LLMs can perform in-context learning without any extra model training or parameter fine-tuning, saving considerable human effort; 2) LLM-based in-context optimization can quickly scale to new tasks or objectives by simply adjusting the prompts, thus enable rapid adaptation to various communication environments; 3) LLMs can provide reasonable explanations for their optimization decisions, helping human understanding complex network systems. 
With these appealing features, LLMs have great potential for handling optimization problems.

On the other hand, the envisioned 6G networks will observe short channel coherence time (due to higher transmission frequencies and narrow beams) and stringent requirements on the quality of network services, calling for faster response time and higher intelligence.  
For example, vehicle-to-infrastructure (V2I) communication is closely related to vehicle's driving behaviours, e.g., frequent acceleration and deceleration can result in extra handovers (HOs) and can even result in V2I connection outages (if not handled on a timely basis)~\cite{10001396}.
% \red{please briefly summarize how AV's behaviour will affect xxx decision of V2I communications such as BS selection etc.}
%
As a result, jointly optimizing V2I communication and autonomous driving (AD) policies is critical. Such integration of V2I and AD also aligns with the 6G usage scenarios defined in International Mobile Telecommunications-2030 (IMT-2030) by the International Telecommunication Union \cite{itu-2030}.  

LLM-based AD has attracted significant research interest, and existing studies have leveraged LLM's reasoning capabilities to enable decision-making \cite{wen2024dilu} and leading human-like AD behaviours \cite{fu2024drive}. 
% While a couple of research works \cite{fu2024drive, wen2024dilu} considered LLMs for AD, the problem of jointly optimizing V2I communication and AD policies has only been solved using traditional RL techniques \cite{10001396,yan2024generalized}. 
% %
% Therefore, the performance gains of using LLM in solving the joint V2I connectivity and AD problem are unexplored.
%
{These existing studies \cite{fu2024drive,wen2024dilu} have explored using LLMs for autonomous driving (AD), but applying LLMs to jointly optimize AD policies and V2I communications is still not investigated. 
While such joint optimization problem has primarily been addressed using RL techniques \cite{10001396,yan2024generalized}, recent progress has witnessed the great potential of LLM techniques.
Consequently, the performance gains of using LLMs in solving joint V2I and AD problems remain unexplored, e.g., higher learning efficiency and explainable decision-making.
In addition, such a joint optimization problem can better reveal the full potential of LLM techniques in addressing highly coupled control problems, i.e., the LLM may need to interact with other agents to make comprehensive decisions.
}
% However, these prior approaches suffer from complex Multi-Objective Markov Decision Processes (MOMDP) designs, which impair training efficiency by including unnecessary state features, such as channel state information, that confuse the AD decision-making process. 
% Notably, \cite{fu2024drive,wen2024dilu} rely solely on the target vehicle's lane position and speed, rarely considering the positions or offsets of surrounding vehicles. Furthermore, these works do not incorporate experience replay, nor do they utilize observation matrices, which are critical for leveraging historical knowledge and enhancing decision-making efficiently. }

{Given the potential of LLMs and the challenges in 6G networks, this work aims to use LLMs to jointly optimize V2I communication and AD policies.} 
%Since most existing LLMs, such as GPT and Llama, are pretrained for general domain purposes, we propose a hybrid LLM-DDQN based iterative solution. 
%
In particular, we first employ LLMs for AD decision-making, aiming to maximize speed, and reduce lane changes and collisions. Using the pre-trained real-world knowledge, LLMs can better understand complicated road environments and quickly make AD decisions. 
We introduce the language-based task description, example design and distance-based selection, enabling LLMs to explore and learn from previous experience. 
{Different from previous studies that employ LLMs only, this work further considers the interactions between LLMs and regular RL agents.}
{Specifically}, the AD decision will be used by the double deep Q-Networks (DDQN) for V2I optimization, which is designed to improve data rates with fewer handovers. The whole problem will be iteratively optimized until convergence. 
{Such a joint optimization problem and iterative optimization scheme can also be generalized to many other real-world scenarios, highlighting the unique contributions of this work.}
{Finally,} our simulation results show that the proposed LLM-based method has higher learning efficiency and faster convergence than baselines. Note that we select DDQN for V2I optimization as the performance of DDQN is well-established and has been demonstrated in many existing studies \cite{10001396,yan2024generalized}.

\begin{figure}
\includegraphics[width=1\linewidth]{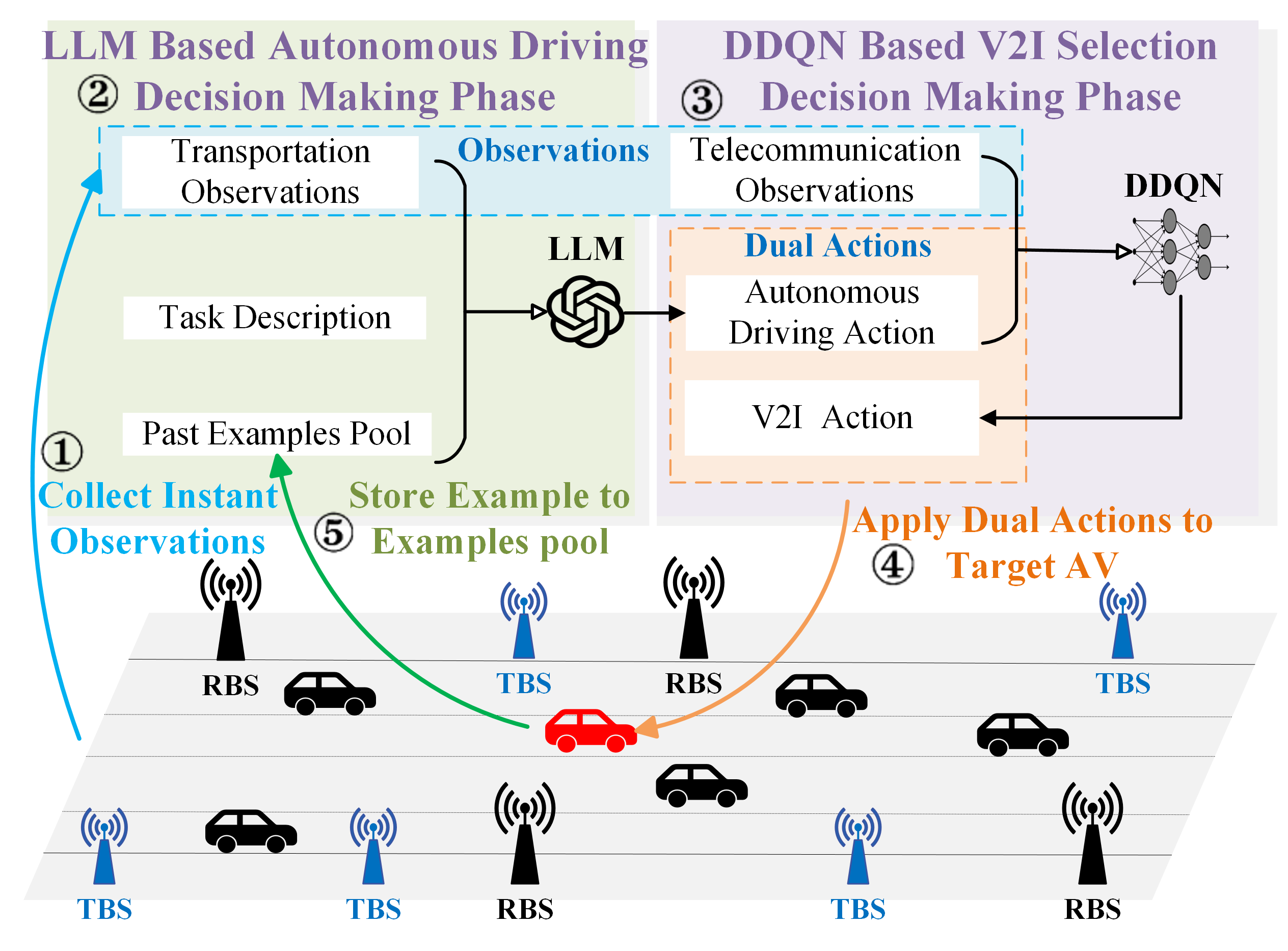}
\caption{Hybrid LLM-DDQN Framework in the RF-THz-Highway Environment}
\label{fig:llm_highway}
\end{figure}

\section{System Model and MDP Definitions}
% $Problem Formulation$
\label{sec:momdp}

This section will introduce the considered system model for AD and V2I connectivity as well as their corresponding Markov decision processes (MDPs). MDP is a very useful approach to define optimization tasks, therefore, we modeled this problem using an MDP.
%{\color{blue} Complicated MDP can affect performance.}

\subsection{V2I and AD System Models}

As shown in Fig.~\ref{fig:llm_highway}, we consider a downlink network comprising $N_R$ RF base stations (RBSs) and $N_T$ THz base stations (TBSs) in a multi-vehicle environment \cite{10001396}. From the V2I perspective, our goal is to maximize data rates and minimize handovers, while from the transportation perspective, we aim to optimize velocity and ensure road safety by minimizing collisions.
We assume $M_1$ autonomous vehicles (AVs), each associated with a single BS (RBS or TBS), receiving data and collecting real-time information from the VNet such as velocity, acceleration, and lane positions of neighboring vehicles. The signal-to-interference-plus-noise ratio (SINR) for the $j$-th AV from RBS $i$ is modeled as \cite{10001396}:
\begin{equation}
\mathrm{SINR}^{\mathrm{RF}}_{ij} = \frac{P_{R}^{\mathrm{tx}} G_{R}^{\mathrm{tx}} G_{R}^{\mathrm{rx}} \left(\frac{c}{4\pi f_{R}} \right)^2 H_{i} }{r_{ij}^{\alpha} (\sigma^2 + I_{R_j})},
\end{equation}
where  $P_{R}^{\mathrm{tx}}, G_{R}^{\mathrm{tx}}, G_{R}^{\mathrm{rx}}, c, f_{R} $, and $\alpha$ denote the transmit power of the RBSs, antenna transmitting gain, antenna receiving gain, speed of light, RF carrier frequency (in GHz) and path-loss exponent, respectively. 
% Note that $r_{ij} = {(d_{ij}^2+h_{ij}^2)}^{1/2},$ where $d_{ij}$ is the 2D distance between the $j$-th AV and $i$-th BS and $h_{ij}$ is the transmit antenna height.
%
Note that $r_{ij} = {(d_{ij}^2+h_{ij}^2)}^{1/2},$ where $d_{ij}$ is the 2D distance between the $j$-th AV and $i$-th BS and $h_{ij}$ is the transmit antenna height.
In addition,
% $$H_i$ is the exponentially distributed channel fading power observed at the {$j$-th} AV from the $i$-th RBS, respectively. 
$H_i$ is the exponentially distributed channel fading power observed at the {$j$-th} AV from the $i$-th RBS,
{$\sigma^2$} is the power of thermal noise at the receiver, $I_{R_j}$ is the cumulative interference at $j$-th AV from the interfering RBSs. 
% We assume all AVs are equipped with a single antenna, and beam alignment techniques ensure the AV's receiving beam aligns with the associated TBS's transmitting beam. The alignment of the main lobes of user and interfering TBSs occurs with probability $q$, and interference is calculated accordingly \cite{mobility}.
Similarly, the SINR for THz networks is modeled as \cite{mobility}:
% \begin{equation}
% \mathrm{SINR}^{\mathrm{THz}}_{ij} = \frac{G_{T}^{\mathrm{tx}} G_{T}^{\mathrm{rx}} \left(\frac{c}{4\pi f_{T}} \right)^2 P_{T}^{\mathrm{tx}} \exp(-K_a(f_T) r_{ij}) r_{ij}^{-2}}{N_{T_j} + I_{T_j}},
% \end{equation}
\begin{equation}
\resizebox{0.9\linewidth}{!}{$
        \mathrm{SINR}^{\mathrm{THz}}_{ij} = \frac{G_{T}^{\mathrm{tx}} G_{T}^{\mathrm{rx}} \left(\frac{c}{4\pi f_{T}} \right)^2 P_{T}^{\mathrm{tx}}\: \mathrm{exp}(-K_a(f_T) r_{ij})r_{ij}^{-2} }{N_{T_j} + I_{T_j}},$
}
\end{equation}
where $G_{T}^{\mathrm{tx}}, G_{T}^{\mathrm{rx}}, P_{T}^{\mathrm{tx}}, f_T$, and $K_a(f_T)$ denote the transmit antenna gain of the TBS, the receiving antenna gain of the TBS, the transmit power of the TBS, the THz carrier frequency, and the molecular absorption coefficient of the transmission medium, respectively.
The alignment of the main lobes of user and interfering TBSs occurs with probability $q$, and interference is calculated accordingly \cite{mobility}. 
The data rate is given by $R_{ij} = W_j \log_2(1 + \mathrm{SINR}_{ij})$, where $W_j$ is the bandwidth of the corresponding RBS or TBS.

In the subsequent subsections, we formulate AD and V2I tasks into two MDPs. 
% LLMs use for AD decision-making and DDQN for V2I optimization.
These MDPs will be further used for LLMs to understand the AD tasks, and DDQN algorithms for V2I decision-making, respectively.

\subsection{Autonomous Driving MDPs}
\label{sec-mdp-transport}

\subsubsection{\textbf{States}}
The transportation state-space consists of kinematics-related features of AVs, represented as  $M_1 \times F_{\mathrm{AD}}$ array, where $F_{\mathrm{AD}} = \{x^j, y^j, v^j, \psi^j\}$ for $j \in [1,M_1]$ describes the $j$-th AV's coordinates $(x^j, y^j)$, velocity $v^j$, and heading $\psi^j$. We model $M_1$ target AVs and $M_2$ surrounding AVs. The aggregated state space $\mathcal{S}^{\mathrm{AD}}$ at time step $t$ is: 
$$
\mathcal{S}^{\mathrm{AD}} =[ \mathbf{x}^j,\mathbf{y}^j,\mathbf{v}^j,\boldsymbol{\psi}^j ]
$$
\subsubsection{\textbf{Actions}}
At each time step $t$, AV $j$ selects an autonomous driving action $a^{\rm AD}_t \in \mathcal{A}_{\rm AD}$, where the driving action space is $\mathcal{A}_{\rm AD} = \{a^{\rm AD}_1, \ldots, a^{\rm AD}_5\}$, representing acceleration, maintaining current lane, deceleration, lane change to left, and lane change to right.

\subsubsection{\textbf{Rewards}}
The transportation reward is defined as:
\begin{equation}
\label{eq:tran_reward}
    r^{\mathrm{j,AD}}_t = c_1 \left( \frac{v^j_t - v_{\min}}{v_{\max} - v_{\min}} \right) - c_2 \delta_2 + c_3 \delta_3 + c_4 \delta_4,
\end{equation}
where $v^j_t$ is the AV’s velocity, and $\delta_2, \delta_3, \delta_4$ represent collision, right-lane, and off-road indicators, respectively. The weights $c_1$ and $c_2$ adjust the AD reward by incorporating a collision penalty. Specifically, $c_1$ scales the reward received when driving at full speed. The reward decreases linearly as the speed decreases, reaching zero at the lowest speed. $c_2$ is the collision penalty, which is significantly larger than the other coefficients. 
To encourage the AV to drive on the rightmost lanes, $c_3$ decreases linearly as the AV deviates from the rightmost lane, reaching zero for other lanes. $c_4$ is the on-road reward factor, which penalizes the AV for driving off the highway.

\subsection{Vehicle-to-Infrastructure MDPs}
\label{sec-mdp-telecom}
We assume each RBS and TBS can support a maximum of $Q_R$ and $Q_T$ AVs, respectively \cite{yan2024generalized}. 
%We assume each AV maintains a list of the top three BSs in terms of achievable data rate, provided $\mathrm{SINR}_{ij}(t) \geq \gamma_{th}$,  where $\gamma_{th}$ represents the desired SINR threshold. 
The connection is satisfactory if the SINR can meet or exceed a threshold $\gamma_{th}$. The number of AVs that each BS can serve at a given time is denoted by $n_i$. As AVs move along the road, they switch BS connections when $\mathrm{SINR}_{ij} < \gamma_{th}$. Frequent HOs will reduce the data rate due to HO latency. To mitigate this, a HO penalty $\mu$ is introduced, with a higher penalty for TBSs and a lower one for RBSs. The \textit{weighted data rate} between BS $i$ and AV $j$  is modeled as follows: 
% $
%     \text{WR}_{ij} = \frac{R_{ij}}{\min \left(Q_i, n_i \right)} (1 - \mu),
% $
\begin{equation}
    \text{WR}_{ij} = \frac{R_{ij}}{\min \left(Q_i, n_i \right)} (1 - \mu),
\end{equation}
where $Q_i = \{ Q_R , Q_T\}$ in terms of  BS $i$ type, $n_i$ is the number of existing users on BS $i$. Given these settings, following is the MDP formulation.

\subsubsection{\textbf{States}}
The telecommunication V2I state space is defined as:
$
\mathcal{S}^{\mathrm{V2I}} = [\mathbf{n}_R^j, \mathbf{n}_T^j ,\mathbf{a}^{j,\mathrm{AD}}],
$
where $\mathbf{n}_R^j$ and $\mathbf{n}_T^j$ represent the numbers of reachable RBSs and TBSs for AV $j$, and $\mathbf{a}^{j,\mathrm{AD}}$ is the transportation action formerly produced by LLM.

\subsubsection{\textbf{Actions}}
AV $j$ selects a V2I action $a^{\rm tele}_t \in \mathcal{A}_{\rm tele}$, where $\mathcal{A}_{\rm tele} = \{a^{\rm tele}_1, a^{\rm tele}_2, a^{\rm tele}_3\}$.  In $a^{\rm tele}_1$, the AV $j$ selects a BS $i$ maximizing the \textit{weighted rate metric} $\text{WR}_{ij}$. In $a^{\rm tele}_2$, the AV selects a BS with maximum $\text{WR}_{ij}$, with $\mu = 0$ if $Q_i \geq n_i(t)$, otherwise selecting the next available BS. In $a^{\rm tele}_3$, the AV connects to the BS with the highest data rate $R_{ij}$.

\subsubsection{\textbf{Rewards}}

Given $\xi^j_t,\text{WR}_{i^*,j,t}$ are the HO rate and the weighted data rate,respectively. The V2I reward is defined as:
\begin{equation}
\label{eq:tele_reward}
    r^{\mathrm{j,V2I}}_t = \text{WR}_{i^*,j,t} \left(1 - \min(1, \xi^j_t)\right),
\end{equation}
% \red{similar to the c values here.}
% { \color{blue}where $c_5$ is used to scalarize the weighted data rate. } \red{Here there is no need to define an extra c5, if it is an independent reward, right? Meanwhile, please change the notations from "tele" to "V2I"}

\section{Hybrid LLM-DDQN based Dual Objective Optimization Algorithm}

% \red{Let's unify the terminology, using "AD" more to instead of transportation, and "V2I" instead of telecom.}

As shown in Fig. \ref{fig:llm_highway}, we propose an iterative optimization framework to integrate LLMs and DDQN to address joint AD and V2I decision-making. The first stage involves utilizing LLMs to make autonomous driving decisions. 
% This stage leverages LLM's capability to infer transportation strategies based on prior examples without the need for extensive fine-tuning or training, significantly accelerating the decision-making process. 
Then the second stage incorporates the DDQN algorithm, focusing on V2I decisions of selecting proper BSs.
In the context of LLM in-context learning \cite{zhou2024large, zhou2024generative}, we integrated the textual context into the formatted model. Our problem is divided into transportation and V2I steps.

\begin{equation}\label{eq-llm}
D_{task}^{\mathrm{AD}} \times \mathcal{E}_{t} \times \mathcal{S}^{\mathrm{AD}}_{t} \times \mathcal{LLM} \Rightarrow a_{t}^{\mathrm{AD}},
\end{equation}    
\begin{equation}\label{eq-dqn}
% D_{task}^{\mathrm{V2I}} \times
\mathcal{S}^{\mathrm{V2I}}_{t} \times a_{t}^{\mathrm{AD}} \times \mathcal{DDQN} \Rightarrow a_{t}^{\mathrm{V2I}},
\end{equation}

% \red{Similar name issues for the notations in the equation.}

Here equation (\ref{eq-llm}) shows the decision-making process of LLMs.
In particular, $D_{task}^{\mathrm{AD}}$ represents the transportation task descriptions, providing fundamental task information to the LLM, i.e., goals and decision variables.
$\mathcal{E}_{t}$ is the set of examples at time $t$, serving as demonstrations for LLMs to learn from, including both positive and negative examples. 
$\mathcal{S}^{\mathrm{AD}}_{t}$ indicates the current environment state which is associated with the target task.
Then $\mathcal{LLM}$ refers to pre-trained LLM models, and $a_{t}^{\mathrm{AD}}$ is the transport decision as we defined in Section \ref{sec-mdp-telecom}.
{
Meanwhile, equation (\ref{eq-dqn}) shows the decision-making of DDQN, including communication network state $\mathcal{S}^{\mathrm{V2I}}_{t}$, DDQN model $\mathcal{DDQN}$ and network decision $a_{t}^{\mathrm{V2I}}$. 
}
It is worth noting that the transportation decision $a_{t}^{\mathrm{AD}}$ is also included in equation (\ref{eq-llm}). It means that the transport decision will affect the performance of V2I systems, and we address this problem using an iterative optimization approach.

\subsection{Language-based Task Description}

This subsection will introduce the defined task description $D_{task}^{\mathrm{AD}}$ in equation (\ref{eq-llm}). Specifically, $D_{task}^{\mathrm{AD}}$ includes the following key components: “\textbf{Task Description}”, “\textbf{Task Goal}”, “\textbf{Task Definition}”, and additional “\textbf{Decision}” criteria. Below, we present a detailed task description to prompt the LLM.

\begin{tcolorbox}[title = {Task Description for AD Policy Selection}]
\label{box1}
\textbf{Task Description}: Assist in driving the ego vehicle on a 4 lanes single direction highway.

\textbf{Task Goal}: 1) Achieve maximum velocity for the ego vehicle while minimizing collisions. 2) Reduce redundant lane changes (\texttt{LANE\_RIGHT}, \texttt{LANE\_LEFT}) unless required for safety. 3) Prefer keeping the vehicle in the right-most lane when safe to do so.

% \textbf{Environment Features}:

\textbf{Task Definition}: Consider the specific Environment features below.
1) \textbf{'x'}: Horizontal offset of the vehicle relative to the ego vehicle along the x-axis.
2) \textbf{'y'}: Vertical offset of the vehicle relative to the ego vehicle along the y-axis.
3) \textbf{'vx'}: Velocity of the vehicle along the x-axis.
4) \textbf{'vy'}: Velocity of the vehicle along the y-axis. A non-zero value indicates lane changes.

Here are some examples of good previous experiences. Consider trying a higher reward action based on these examples: \{\textit{Good Examples Set}\}

Here are some examples of poor previous experiences. It is suggested to avoid selecting these actions based on these examples: \{\textit{Bad Examples Set}\} 

\textbf{Decisions}: Choose one action from \texttt{FASTER}, \texttt{SLOWER}, \texttt{LANE\_RIGHT}, \texttt{LANE\_LEFT}, or \texttt{IDLE}.

% \textbf{Decision}: Please provide only the chosen action in the response.
\end{tcolorbox}

Firstly, the \textbf{Task Description} defines the task as "\textit{Assist in driving the ego vehicle on a 4-lane single-direction highway.}" The \textbf{Task Goal} is to achieve three objectives for optimizing autonomous driving.
The \textbf{Task Definition} introduces the environment states that the agent AV needs to evaluate. It includes four critical features defined in equation (\ref{eq:tran_reward}) for the target AV and four surrounding AVs. 
%We discretize the observation information for these $M_1$ AVs and  compute the post-processing matrix, which has dimensions $M_1 \times N_f$, where $N_f$ is the number of features in observation states.  
Then, we include good examples set $\mathcal{E}^{\mathrm{good}}_{t}$ and bad examples set $\mathcal{E}^{\mathrm{bad}}_{t}$ by "\textit{Here are some examples of good/bad previous experiences...}". It aims to provide relevant previous experiences to support LLMs in addressing unseen environments. 
Finally, we set extra reply rules such as "\textit{Choose one action from ...}", guiding the LLM to focus on the decision-making process.
Such a task description provides a template to define optimization tasks using formatted natural language, avoiding the complexity of dedicated optimization model design.

 % efficient training and
% and training.  
%It reduces confusion by avoiding self-contradictory examples and only providing relevant past experience examples, thereby optimizing the decision-making process.
\label{subsection:prompt}

\subsection{Example Design and Distance-based Selection}
\label{subsection:example_design}
The above discussions have shown the crucial importance of demonstrations, which are inspired by the well-known “experience pool” in DRL algorithms. 
There are two major challenges in selecting examples. First, we must provide accurate and unbiased examples to ensure accuracy. Second, we cannot send infinite examples to the LLM due to token number constraints. Furthermore, in our joint optimization problem, the ideal decision for autonomous driving will enhance V2I training performance in the next phase. 
Since the features defined in $\mathcal{S}^{\mathrm{AD}}$ have an infinite number of examples, identifying useful examples remains challenging. The transportation example 
$E_t \in \mathcal{E}$ is defined by:
\begin{equation}
    E_t = \{ \mathcal{S}^{\mathrm{AD}}_t , a^{\mathrm{AD}}_t, r^{\mathrm{ego,AD}}_t \}, 
\end{equation}
We classify the examples as good or bad based on collisions. After action $a_{t}^{\mathrm{AD}}$ is applied to the target AV, if the AV truncates or collides, we add this example $E_t$ to the bad example pool $\mathcal{E}^{\mathrm{bad}}_{t} = \mathcal{E}^{\mathrm{bad}}_{t-1} \cup E_t$, and the good example pool remains the same as the last time step. Otherwise, we add this example to the good example pool $\mathcal{E}^{\mathrm{good}}_{t} = \mathcal{E}^{\mathrm{good}}_{t-1} \cup E_t$, and the bad example pool remains the same as the last time step.

We select the top-$K$ examples with the closest Euclidean distance.
% \red{Similarly, use "top-$K$" to replace top 5.}
%
After accumulating good and bad examples, we need to prioritize the relevant appropriate examples. In this work, we select the top-$K$ relevant examples per category (good or bad) via Euclidean distance before forwarding them to LLM agents. Considering we want to filter the relevant good examples from $\mathcal{E}^{\mathrm{good}}_{t-1}$ for the current state $\mathcal{S}^{\mathrm{AD}}_t$, which dimension is $ M_1 \times N_f$. We flat this 2D state to a scalar series $\{\left( \mathbf{x}^1_t,\mathbf{y}^1_t,\mathbf{v}^1_t,\boldsymbol{\psi}^1_t \right), \dots , \left( \mathbf{x}^{M_1}_t,\mathbf{y}^{M_1}_t,\mathbf{v}^{M_1}_t,\boldsymbol{\psi}^{M_1}_t \right) \}$, {euclidean distance $d$ is defined as:
$    d  =  \sum_{j=1}^{M_1} \left\lVert \left( \mathbf{x}^j_t, \mathbf{y}^j_t, \mathbf{v}^j_t, \boldsymbol{\psi}^j_t \right) - \left( \mathbf{x}^j_{E_t}, \mathbf{y}^j_{E_t}, \mathbf{v}^j_{E_t}, \boldsymbol{\psi}^j_{E_t} \right) \right\rVert,$

For each feature pair between the instant observation at time step $t$ and an example in the $E_t \in \mathcal{E}_{t-1}$. We calculate the euclidean distance between the closest $M_1$ AVs. $d$ is smaller while $E_t$ is similar to current observation. 
This approach ensures we collect the $K$ closest relevant past examples and sort them based on the highest reward $r^{\mathrm{ego,AD}}_t$ 
{to ensure real-time decision-making}.

\begin{figure*}
\includegraphics[width=1\linewidth,height=7.5cm]{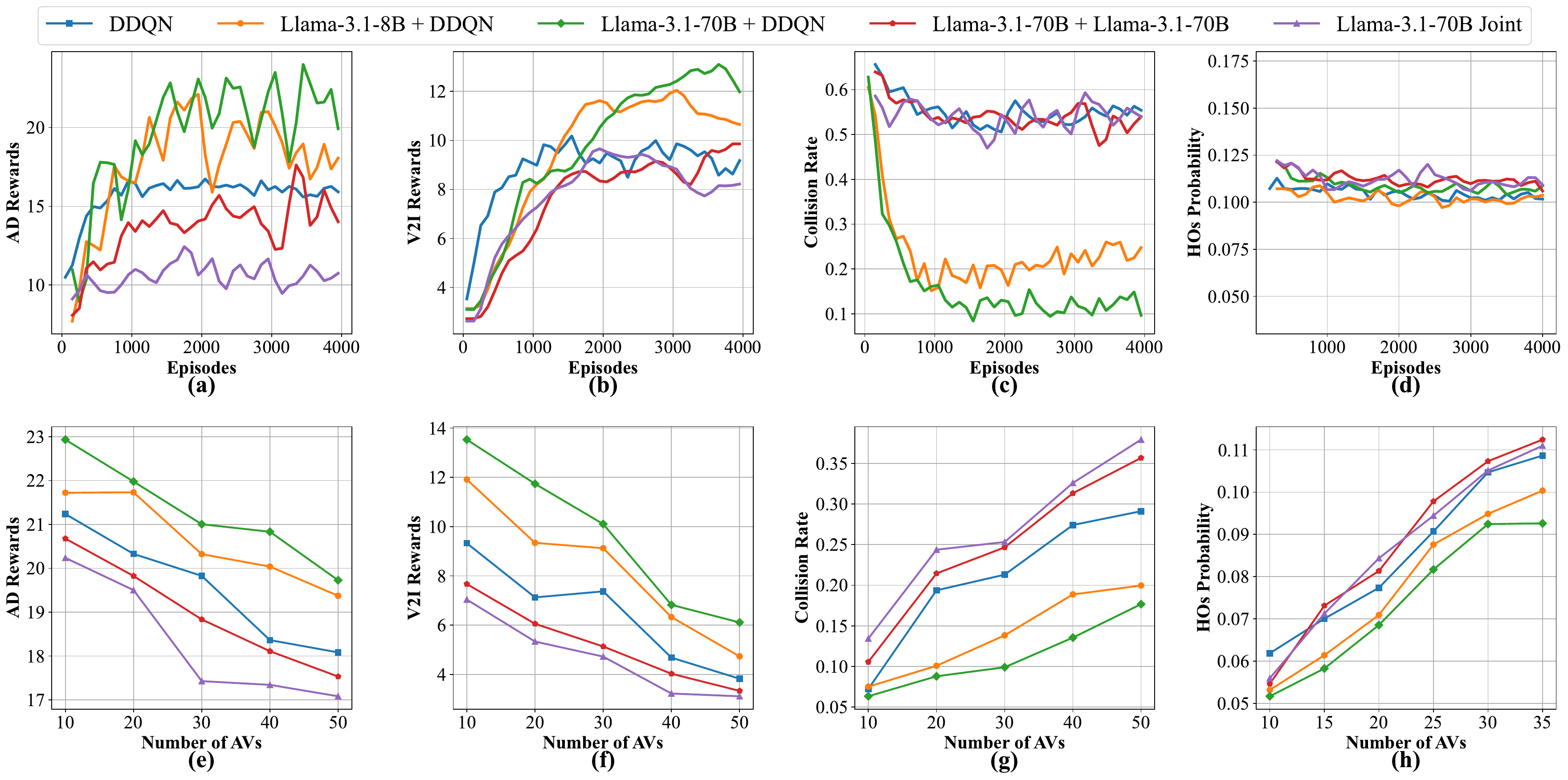} %\hspace
\centering
\caption{Simulation results and comparisons}
\vspace{-5mm}
\label{fig:llama_envelope_ddqn_training}
\end{figure*}

\subsection{DDQN Based V2I  Action Selection}
\label{subsection:ddqn}

As depicted in Fig.~\ref{fig:llm_highway}, LLMs first determine the AD action ${a}^{\mathrm{AD}}_j$, while V2I observations ${n}_R^j$ and ${n}_T^j$ are collected. The closest $M_1 - 1$ AVs with these features form an MDP, which DDQN possesses to compute the V2I action ${a}^{\mathrm{V2I}}_j$. 
DDQN uses a neural network, parameterized by $\boldsymbol{\theta}_t$, to approximate the $Q$-value function for state-action pairs, $Q(\mathcal{S}^{\mathrm{V2I}}_t , a^{\mathrm{AD}}_t , a^{\mathrm{V2I}}_t; \boldsymbol{\theta}_t)$. A separate target network with parameters $\boldsymbol{\theta}_t'$ ensures stability by providing $Q$-value estimates for the target. The evaluation network minimizes the mean squared error (MSE) loss $\mathcal{L}(\boldsymbol{\theta}_t)$ between the predicted $Q$-values and the target $Q$-values $\hat{Q}(\mathcal{S}^{\mathrm{V2I}}_t , a^{\mathrm{AD}}_t , a^{\mathrm{V2I}}_t; \boldsymbol{\theta}_t')$, following the Bellman equation as,

\begin{equation}
    \resizebox{0.9\hsize}{!}{%
        $\begin{aligned}
        \mathcal{L}(\boldsymbol{\theta}_t) = \mathbb{E}_{(s, a, r, s')} \Bigg[ \Big( r^{\mathrm{j,tele}}_t + \gamma Q\big(&\mathcal{S}^{\mathrm{V2I}}_{t+1},  \\ 
        \arg\max_{{a^{\mathrm{V2I}}_{t+1}}'} Q(\mathcal{S}^{\mathrm{V2I}}_{t+1}, {{a^{\mathrm{V2I}}_{t+1}}'}; \boldsymbol{\theta}_t); \boldsymbol{\theta}_t' \big)
        & - Q(\mathcal{S}^{\mathrm{V2I}}_t , a^{\mathrm{AD}}_t , a^{\mathrm{V2I}}_t; \boldsymbol{\theta}_t) \Big)^2 \Bigg]
        \end{aligned}$%
    }
\end{equation}

\subsection{Hybrid LLM-DDQN Learning and Convergence Analysis}
Our proposed hybrid LLM-DDQN scheme consists of the following steps:
\textbf{1) Collecting observations}:
The observation pool collects both V2I and AD data from the ego AV $j$ and the surrounding $M_1$ AVs. The observation will be separated into Transportation observations and V2I observations.
\textbf{2) LLM-based AD decision-making}: 
Using the transportation observations, relevant past examples are computed as described in Section \ref{subsection:example_design}. This information is then incorporated into a task-oriented prompt, as outlined in Section \ref{subsection:prompt}. The prompt is sent to the LLM, which returns an AD decision for the current time step.
\textbf{3) DDQN-based V2I decision-making}:
Using the V2I observations from step 1 and the AD decision from step 2, the V2I action is determined via DDQN as in Section \ref{subsection:ddqn}.
\textbf{4) Applying actions to target AVs}:
Both AD and V2I actions from steps 1 and 2, is applied to the target AV for implementation.
\textbf{5) Storing examples in the example pool}:
After the AV executes its actions, if the AV terminates unexpectedly, the example is stored in the $\mathcal{E}^{\mathrm{bad}}$; otherwise, it is stored in the $\mathcal{E}^{\mathrm{good}}$ as explained in Section \ref{subsection:example_design}.

Finally, we will analyze the convergence of the proposed hybrid LLM-DDQN scheme. It is worth noting that LLM-based AD decisions will serve as part of the V2I states as defined in Section \ref{sec-mdp-telecom}. It means that AD decisions can be considered as environment states for DDQN-based V2I systems. 
Therefore, for the AD side, LLM will produce a stable policy when enough examples are collected, which is similar to DDQN algorithms with experience pools.    
For the V2I side, since AD decisions can be considered as external states, the overall convergence will still hold.

\section{Simulation and Performance Evaluation}

We assume a total of 21 AVs traveling on a 3km single-direction, 4-lane highway.
Along both sides of the highway, 5 RBSs and 20 TBSs are randomly positioned. 
This work includes two LLMs for AD decision-making:
1) \textbf{Llama3.1-8B + DDQN}: Llama3.1-8B is a small-scale model with 8 billion parameters, which is suitable for deployment at the network edge.
2) \textbf{Llama3.1-70B + DDQN}: Llama3.1-70B is the latest large-scale LLM model with 70 billion parameters, which has more powerful reasoning capabilities. 
3) \textbf{Llama3.1-70B + Llama3.1-70B}: Employing two sequential Llama3.1 agents to perform AD and V2I actions simultaneously.  
4) \textbf{Llama3.1-70B Joint}: Employing single Llama3.1 agents to generate Joint AD and V2I actions directly.
We also include a \textbf{DDQN baseline algorithm}, which employs DDQN for joint optimization of V2I and AD \cite{yan2024generalized}.

{
Table~\ref{tab:evaluation_models} evaluates the AD rewards, V2I rewards, and total rewards as defined in \cite{yan2024generalized} for ChatGPT-3.5, Llama3.1-8B, and Llama3.1-70B. The results highlight the superior performance of Llama3.1-70B and Llama3.1-8B compared to ChatGPT-3.5. These models consistently achieve higher rewards across all metrics, demonstrating their effectiveness in optimizing AD and V2I performance.
We also found that increasing the number of AVs slightly reduces performance. This is due to the increased collision probability and intense channel competition in the simulation environment as the number of AVs increases.}

Then Fig. \ref{fig:llama_envelope_ddqn_training} presents the simulation results and comparisons. The performance metrics include the AD reward and V2I reward, as defined in (\ref{eq:tran_reward}) and (\ref{eq:tele_reward}), along with the collision rate and HO probability.
In particular, Fig. \ref{fig:llama_envelope_ddqn_training}(a) and (b) demonstrate that all LLM-based models show an improvement in both AD and V2I rewards, outperforming the DDQN baseline. 
{Especially, Fig.~\ref{fig:llama_envelope_ddqn_training}(c) shows that the hybrid LLM-DDQN method achieves a lower collision rate than the pure DDQN method. It highlights LLM's capabilities in efficient and high-quality exploration for critical missions such as AD.
Additionally, these results highlight the ability of hybrid LLM-DDQN models to learn effectively from past examples, and follow language-based instructions to enhance the overall performance on target tasks with successful collaboration.}

{
\begin{table}[!t]\centering \footnotesize
    \vspace{-0.1cm}
    \renewcommand{\arraystretch}{1.1}
    \setlength\tabcolsep{0.8em}
        \resizebox{\linewidth}{!}{%  \textwidth
    \begin{tabular}{cccccc}
    \toprule
    \toprule
     \multicolumn{1}{c}{\shortstack{Desire Velocity}} & \multicolumn{1}{c}{\multirow{2}{*}{Model}} & \multicolumn{3}{c}{Number of AVs $M$} \\
     \cmidrule(lr){3-5}
     \multicolumn{1}{c}{(m/s)} & \multicolumn{1}{c}{} & \multicolumn{1}{c}{$M$=10} & \multicolumn{1}{c}{$M$=20} & \multicolumn{1}{c}{$M$=30} \\
    \midrule
    \multirow{3}{*}{15} & ChatGPT-3.5 & 24.72/11.25/35.97 & 24.28/10.41/34.69 & 22.26/8.43/30.69 \\
    & Llama3.1-8B & 25.38/13.66/39.04 & 24.99/10.45/35.44 & 22.17/9.09/31.26 \\
    & Llama3.1-70B & \textbf{26.12/15.21/41.33} & \textbf{25.41/12.26/37.67} & \textbf{23.49/10.28/33.77} \\
    \midrule
    \multirow{3}{*}{20} & ChatGPT-3.5 & 21.18/9.42/30.60 & 20.44/7.08/27.52 & 19.91/7.55/27.46 \\
    & Llama3.1-8B & 21.76/11.99/33.75 & 21.78/9.37/31.15 & 20.36/9.21/29.57 \\
    & Llama3.1-70B & \textbf{22.88/13.78/36.66} & \textbf{21.99/11.36/33.35} & \textbf{21.02/10.15/31.17} \\
    \midrule
    \multirow{3}{*}{25} & ChatGPT-3.5 & 17.72/8.32/26.04 & 16.42/6.41/22.83 & 15.83/5.89/21.72 \\
    & Llama3.1-8B & \textbf{18.89/8.70/27.59} & 16.59/6.38/22.97 & 16.22/5.94/22.16 \\
    & Llama3.1-70B & 18.34/8.45/26.79 & \textbf{16.78/6.46/23.24} & \textbf{16.38/6.03/22.41} \\
    \midrule
    \multirow{3}{*}{30} & ChatGPT-3.5 & 16.51/7.31/23.82 & 16.44/5.77/22.21 & 15.38/5.71/21.09 \\
    & Llama3.1-8B & 16.93/7.65/24.58 & 16.81/5.92/22.73 & \textbf{16.51/5.82/22.33} \\
    & Llama3.1-70B & \textbf{17.42/7.92/25.34} & \textbf{17.04/6.12/23.16} & 15.42/5.81/21.23 \\
    \bottomrule
    \bottomrule
    \end{tabular}
    }
    \caption{Evaluation performance of GPT-3.5, Llama3.1-8B, and Llama3.1-70B (AD rewards/ V2I rewards/ Total Rewards).}
    \label{tab:evaluation_models}
    \vspace{-0.2cm}
\end{table}
}

Fig. \ref{fig:llama_envelope_ddqn_training}(e) to (h) depict the performance variation as the number of AVs increases. Each data is captured by using the average performance of the last 200 episodes .
% \textbf{Impact of the Number of AVs:}
As shown, increasing the number of AVs results in more congested highway scenarios, leading to more competitive resource allocation. This, in turn, results in higher HOs among the travelling vehicles. 
{However, our proposed hybrid Llama+DDQN method still outperforms conventional DDQN approaches with higher average reward and lower collision rate.}
Finally, note that the LLM performance is closely related to its model size and designs. Specifically, Llama-3.1-70B outperforms Llama-3.1-8B because a larger model size usually indicates better reasoning and comprehension capabilities. 
Dual-agent or joint decision-making using LLMs does not perform well, as multi-task and multi-objective problems remain challenging for LLMs. These models struggle to learn effectively in complex scenarios with numerous constraints.

\section{Conclusion}

LLMs have great potential for 6G networks, and this work proposed a hybrid LLM-DDQN scheme for joint V2I and AD optimization. 
It explores natural language-based optimization and network management, revealing the potential of LLM techniques. 
The simulations demonstrate that our proposed method can achieve faster convergence and higher learning efficiency than existing DDQN techniques.

% that's all folks
\bibliographystyle{IEEEtran}
\bibliography{main.bib}
\end{document}